# ProjE: Embedding Projection for Knowledge Graph Completion

Baoxu Shi[1] and Tim Weninger[1]

[1]*Department of Computer Science and Engineering, University of Notre Dame*

**Abstract**

With the large volume of new information created every day, determining the validity of information in a knowledge graph and filling in its missing parts are crucial tasks for many researchers and practitioners. To address this challenge, a number of knowledge graph completion methods have been developed using low-dimensional graph embeddings. Although researchers continue to improve these models using an increasingly complex feature space, we show that simple changes in the architecture of the underlying model can outperform state-of-the-art models without the need for complex feature engineering. In this work, we present a shared variable neural network model called ProjE that fills-in missing information in a knowledge graph by learning joint embeddings of the knowledge graph's entities and edges, and through subtle, but important, changes to the standard loss function. In doing so, ProjE has a parameter size that is smaller than 11 out of 15 existing methods while performing 37% better than the current-best method on standard datasets. We also show, via a new fact checking task, that ProjE is capable of accurately determining the veracity of many declarative statements.

Knowledge Graphs (KGs) have become a crucial resource for many tasks in machine learning, data mining, and artificial intelligence applications including question answering [34], entity disambiguation [7], named entity linking [14], fact checking [32], and link prediction [28] to name a few. In our view, KGs are an example of a heterogeneous information network containing entity-nodes and relationship-edges corresponding to RDF-style triples $\langle h, r, t \rangle$ where $h$ represents a head entity, and $r$ is a relationship that connects $h$ to a tail entity $t$.

KGs are widely used for many practical tasks, however, their correctness and completeness are not guaranteed. Therefore, it is necessary to develop *knowledge graph completion* (KGC) methods to find missing or errant relationships with the goal of improving the general quality of KGs, which, in turn, can be used to improve or create interesting downstream applications.

The KGC task can be divided into two non-mutually exclusive sub-tasks: (i) entity prediction and (ii) relationship prediction. The *entity* prediction task takes a partial triple $\langle h, r, ? \rangle$ as input and produces a ranked list of candidate entities as output:

**Definition 1.** *(Entity Ranking Problem) Given a Knowledge Graph $\mathcal{G} = \{\mathbf{E}, \mathbf{R}\}$ and an input triple $\langle h, r, ? \rangle$, the* entity ranking problem *attempts to find the optimal ordered list such that $\forall e_j \forall e_i ((e_j \in \mathbf{E}_- \land e_i \in \mathbf{E}_+) \rightarrow e_i \prec e_j)$, where $\mathbf{E}_+ = \{e \in \{e_1, e_2, \ldots, e_l\} | \langle h, r, e \rangle \in \mathcal{G}\}$ and $\mathbf{E}_- = \{e \in \{e_{l+1}, e_{l+2}, \ldots, e_{|E|}\} | \langle h, r, e \rangle \notin \mathcal{G}\}$.*

Distinguishing between head and tail-entities is usually arbitrary, so we can easily substitute $\langle h, r, ? \rangle$ for $\langle ?, r, t \rangle$.

The *relationship* prediction task aims to find a ranked list of relationships that connect a head-entity with a tail-entity, *i.e.*, $\langle h, ?, t \rangle$. When discussing the details of the present work, we focus specifically on the entity prediction task; however, it is straightforward to adapt the methodology to the relationship prediction task by changing the input.

A number of KGC algorithms have been developed in recent years, and the most successful models all have one thing in common: they use low-dimensional embedding vectors to represent entities and relationships. Many embedding models, *e.g.*, Unstructured [3], TransE [4], TransH [35], and TransR [25], use a margin-based pairwise ranking loss function, which measures the score of each possible result as the $L_n$-distance between $h + r$ and $t$. In these models the loss functions are all the same, so models differ in how they transform the entity embeddings $h$ and $t$ with respect to the relationship embeddings $r$. Instead of simply adding $h + r$, more expressive *combination operators* are learned by Knowledge Vault [8] and HolE [29] in order to predict the existence of $\langle h, r, t \rangle$ in the KG.

Other models, such as the Neural Tensor Network (NTN) [33] and the Compositional Vector Space Model (CVSM) [27], incorporate a multilayer neural network solution into the existing models. Unfortunately, due to their extremely large parameter size, these models either (i) do not scale well or (2) consider only a single relationship at a time [10] thereby limiting their usefulness on large, real-world KGs.

Despite their large model size, the aforementioned methods only use singleton triples, *i.e.*, length-1 paths in the KG. PTransE [24] and RTransE [10] employ extended path information from 2 and 3-hop trails over the knowledge graph. These extended models achieve excellent performance due to the richness of the input data; unfortunately, their model-size grows exponentially as the path-length increases, which further exacerbates the scalability issues associated with the already high number of parameters of the underlying-models.

Another curious finding is that some of the existing models are not *self-contained models*, *i.e.*, they require pre-trained KG embeddings (RTransE, CVSM), pre-selected paths (PTransE, RTransE), or pre-computed content embeddings of each node (DKRL [36]) before their model training can even begin. TransR and TransH are self-contained models, but their experiments only report results using pre-trained TransE embeddings as input.

With these considerations in mind, in the present work we rethink some of the basic decisions made by previous models to create a *projection embedding* model (ProjE) for KGC. ProjE has four parts that distinguish it from the related work:

1. Instead of measuring the distance between input triple $\langle h, r, ? \rangle$ and entity candidates on a unified or a relationship-specific plane, we choose to project the entity candidates onto a target vector representing the input data.
2. Unlike existing models that use transformation matrices, we combine the embedding vectors representing the input data into a target vector using a learnable combination operator. This avoids the addition of a large number of transformation matrices by reusing the entity-embeddings.
3. Rather than optimizing the margin-based pairwise ranking loss, we optimize a ranking loss of the list of candidate-entities (or relationships) collectively. We further use candidate sampling to handle very large data sets.
4. Unlike many of the related models that require pre-trained data from prerequisite models or explore expensive multi-hop paths through the knowledge graph, ProjE is a self-contained model over length-1 edges.

# 1 Related Work

A variety of low-dimensional representation-based methods have been developed to work on the KGC task. These methods usually learn continuous, low-dimensional vector representations (*i.e.*, embeddings) for entities $\mathbf{W}^E$ and relationships $\mathbf{W}^R$ by minimizing a margin-based pairwise ranking loss [24].

The most widely used embedding model in this category is TransE [4], which views relationships as translations from a head entity to a tail entity on the same low-dimensional plane. The energy function of TransE is defined as

$$E(h, r, t) = \| \mathbf{h} + \mathbf{r} - \mathbf{t} \|_{L_n}, \qquad (1)$$

which measures the $L_n$-distance between a translated head entity $\mathbf{h} + \mathbf{r}$ and some tail entity $\mathbf{t}$. The Unstructured model [3] is a special case of TransE where $\mathbf{r} = \mathbf{0}$ for all relationships.

Based on the initial idea of treating two entities as a translation of one another (via their relationship) in the same embedding plane, several models have been introduced to improve the initial TransE model. The newest contributions in this line of work focus primarily on the changes in how the embedding planes are computed and/or how the embeddings are combined. For example, the entity translations in TransH [35] are computed on a hyperplane that is perpendicular to the relationship embedding. In TransR [25] the entities and relationships are embedded on separate planes and then the entity-vectors are translated to the relationship's plane. Structured Embedding (SE) [5] creates two translation matrices for each relationship and applies them to head and tail entities separately. Knowledge Vault [8] and HolE [29], on the other hand, focus on learning a new combination operator instead of simply adding two entity embeddings element-wise.

The aforementioned models are all geared toward link prediction in KGs, and they all minimize a margin-based pairwise ranking loss function $\mathcal{L}$ over the training data $\mathbf{S}$:

$$\mathcal{L}(\mathbf{S}) = \Sigma_{(h,r,t) \in \mathbf{S}}[\gamma + E(h, r, t) - E(h', r', t')]_+, \qquad (2)$$

where $E(h, r, t)$ is the energy function of each model, $\gamma$ is the margin, and $(h', r', t')$ denotes some "corrupted" triple which does not exist in $\mathbf{S}$. Unlike aforementioned models that focus on different $E(h, r, t)$, TransA [19] introduces an adaptive local margin approach that determines $\gamma$ by a closed set of entity candidates. Other similar models include RESCAL [30], Semantic Matching Energy (SME) [3], and the Latent Factor Model (LFM) [18].

The Neural Tensor Network (NTN) model [33] is an exception to the basic energy function in Eq. 1. Instead, NTN uses an energy function

$$E(h, r, t) = \mathbf{u}_r^T f(\mathbf{h}^T \mathbf{W}_r \mathbf{t} + W_{rh}\mathbf{h} + W_{rt}\mathbf{t} + \mathbf{b}_r), \qquad (3)$$

where $\mathbf{u}_r$, $\mathbf{W}_r$, $W_{rh}$, and $W_{rt}$ are all relationship-specific variables. As a result, the number of parameters in NTN is significantly larger than other methods. This makes NTN unsuitable for networks with even a moderate number of relationships.

So far, the related models have only considered triples that contain a single relationship. More complex models have been introduced to leverage path and content information in KGs. For instance, the Compositional Vector Space Model (CVSM) [27] composes a sequence of relationship embeddings into a single path embedding using a Recurrent Neural Network (RNN). However, this has two disadvantages: (i) CVSM needs pre-trained relationship embeddings as input, and (ii) each CVSM is specifically trained for only a single relationship type. This makes CVSM perform well in specific tasks, but unsuitable for generalized entity and relationship prediction tasks. RTransE [10] solves the relationship-specific problem in CVSM by using entity and relationship embeddings learned from TransE. However, it is hard to compare RTransE with existing methods because it requires unambiguous, pre-selected paths as inputs called quadruples $\langle h, r_1, r_2, t \rangle$ further complicating the model. DKRL, like NTN, uses word embeddings of entity-content in addition to multi-hop paths, but relies on the machinery of a Convolution Neural Network (CNN) to learn entity and relationship embeddings.

PTransE [24] is another path-based method that uses path information in its energy function. Simply put, PTransE doubles the number of edges in the KG by creating reverse relationships for every existing relationship in the KG. Then PTransE uses PCRA [37] to select input paths within a given length constraint.

Table 1 shows a breakdown of the parameter complexity of each model. As is typical, we find that more complex models achieve better prediction accuracy, but are also more difficult to train and have trouble scaling. The proposed method, ProjE, has a number of parameters that is smaller than 11 out of 15 methods and does not require any prerequisite training.

## 2 Methodology

The present work views the KGC problem as a ranking task and optimizes the collective scores of the list of candidate entities. Because we want to optimize the ordering of candidate entities collectively, we need to project the candidate entities onto the same embedding vector. For this task we learn a combination operator that creates a target vector from the input data. Then, the candidate entities are each projected onto the same target vector thereby revealing the candidate's similarity score as a scalar.

In this section we describe the ProjE architecture, followed by two proposed variants, their loss functions, and our choice of candidate sampling method. In the experiments section we demonstrate that ProjE outperforms all existing methods despite having a relatively small parameter space. A detailed algorithm description can be found in the Supplementary Material.

Table 1: Parameter size and prerequisites of KGC models in increasing order. ProjE, ranked 5[th], is highlighted. $n_e, n_r, n_w, k$ are the number of entities, relationships, words, and embedding size in the KG respectively. $z$ is the hidden layer size. $q^\dagger$ represents the number of RNN parameters in RTransE; this value is not specified, but should be $8k^2$ if a normal LSTM is used.

| Model | Parameters | Prerequisites |
|---|---|---|
| Unstructured | $n_e k$ | - |
| TransE | $n_e k + n_r k$ | - |
| HolE | $n_e k + n_r k$ | - |
| PTransE | $n_e k + n_r k$ | PCRA |
| **ProjE** | $n_e k + n_r k + 5k$ | - |
| CVSM | $n_e k + n_r k + 2k^2$ | Word2vec |
| SME (linear) | $n_e k + n_r k + 4k^2$ | - |
| RTransE | $n_e k + n_r k + q^\dagger$ | TransE, PCRW |
| LFM | $n_e k + n_r k + 10k^2$ | - |
| SME (bilinear) | $n_e k + n_r k + 2k^3$ | - |
| TransH | $n_e k + 2 n_r k$ | - |
| RESCAL | $n_e k + n_r k^2$ | - |
| SE | $n_e k + 2 n_r k^2$ | - |
| TransR | $n_e k + n_r(k + k^2)$ | - |
| DKRL | $n_e k + n_r k + n_w k + 2zk$ | TransE, Word2vec |
| NTN | $n_e k + n_r(zk^2 + 2zk + 2z)$ | - |

(increasing model size)

## 2.1 Model Architecture

The main insight in the development of ProjE is as follows: given two input embeddings, we view the prediction task as ranking problem where the top-ranked candidates are the correct entities. To generate this ordered list, we project each of the candidates onto a target vector defined by two input embeddings through a combination operator.

Existing models, such as Knowledge Vault, HolE, and NTN, define specific matrix combination operators that combine entities and/or relationships. In common practice, these matrices are expected to be sparse. Because we believe it is unnecessary to have interactions among different feature dimensions at this early stage, we constraint our matrices to be diagonal, which are inherently sparse. The combination operator is therefore defined as

$$\mathbf{e} \oplus \mathbf{r} = \mathbf{D}_e \mathbf{e} + \mathbf{D}_r \mathbf{r} + \mathbf{b}_c, \qquad (4)$$

where $\mathbf{D}_e$ and $\mathbf{D}_r$ are $k \times k$ diagonal matrices which serve as global entity and relationship weights respectively, and $\mathbf{b}_c \in \mathcal{R}^k$ is the combination bias.

Using this combination operator, we can define the embedding projection function as

$$h(\mathbf{e}, \mathbf{r}) = g(\mathbf{W}^c f(\mathbf{e} \oplus \mathbf{r}) + b_p), \qquad (5)$$

where $f$ and $g$ are activation functions that we define later, $\mathbf{W}^c \in \mathcal{R}^{s \times k}$ is the candidate-entity matrix, $b_p$ is the projection bias, and $s$ is the number of candidate-entities. $h(\mathbf{e}, \mathbf{r})$ represents the ranking score vector, where each element represents the similarity between some candidate entity in $\mathbf{W}^c$ and the combined input embedding $\mathbf{e} \oplus \mathbf{r}$.

Although $s$ is relatively large, due to the use of shared variables, $\mathbf{W}^c$ is the candidate-entity matrix that contains $s$ rows that exist in the entity embedding matrix $\mathbf{W}^E$. Simply put, $\mathbf{W}^c$ does not introduce any new variables into the model. Therefore, compared to simple models like TransE, ProjE only increases the number of parameters by $5k + 1$, where $1$, $4k$, and $k$ are introduced as the projection bias, combination weights, and combination bias respectively. Later we show that by changing different activation functions, ProjE can be either a pointwise ranking model or a listwise ranking model.

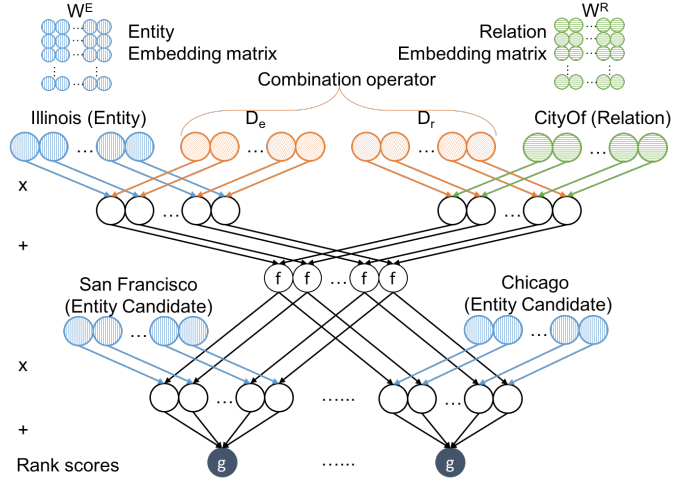

**Figure 1:** ProjE architecture for entity prediction with example input $\langle ?, \textsf{CityOf}, \textsf{Illinois} \rangle$ and two candidates. ProjE represents a two-layer neural network with a combination layer, and a projection (*i.e.*, output) layer. This figure is best viewed in color.

ProjE can be viewed as a neural network with a combination layer and a projection (*i.e.*, output) layer. Figure 1 illustrates this architecture by way of an example. Given a tail entity Illinois and a relationship CityOf, our task is to calculate the scores of each head entity. The blue nodes are row vectors from the entity embedding matrix $\mathbf{W}^E$, and the green nodes are row vectors from the relationship embedding matrix $\mathbf{W}^R$; the orange nodes are the combination operators as diagonal matrices. For clarity we only illustrate two candidates in Fig. 1, however $\mathbf{W}^c$ may contain an arbitrary number of candidate-entities.

The next step is to define the loss functions used in ProjE.

## 2.2 Ranking Method and Loss Function

As defined in Defn. 1, we view the KGC problem as a ranking task where all positive candidates precede all negative candidates and train our model accordingly. Typically there are two ways to obtain such an ordering: with either 1) the pointwise method, or 2) the listwise method [31]. Although most existing KGC models, including TransE, TransR, TransH, and HolE use a pairwise ranking loss function during training, their ranking score is calculated independently in what is essentially a pointwise method when deployed. Based on the architecture we described in previous section, we propose two methods: 1) ProjE_pointwise, and 2) ProjE_listwise through the use of different activation functions for $g(\cdot)$ and $f(\cdot)$ in Eq. 5.

First we describe the **ProjE_pointwise** ranking method. Because the relative order inside each entity set does not affect the prediction power, we can create a binary label vector in which all entities in $\mathbf{E}_-$ have a score of 0, and all entities in $\mathbf{E}_+$ have a score of 1. Because we maximize the likelihood between the ranking score vector $h(\mathbf{e}, \mathbf{r})$ and the binary label vector, it is intuitive to view this task as a multi-class classification problem. Therefore, the loss function of ProjE_pointwise can be defined in a familiar way:

$$\mathcal{L}(\mathbf{e}, \mathbf{r}, \mathbf{y}) = - \sum_{i \in \{i | y_i = 1\}} \log(h(\mathbf{e}, \mathbf{r})_i) \\ - \sum_m \mathbb{E}_{j \sim P_y} \log(1 - h(\mathbf{e}, \mathbf{r})_j), \quad (6)$$

where $\mathbf{e}$ and $\mathbf{r}$ are the input embedding vectors of a training instance in $\mathbf{S}$, $\mathbf{y} \in \mathcal{R}^s$ is a binary label vector where $y_i = 1$ means candidate $i$ represents a positive label, $m$ is the number of negative samples drawn from a negative candidate distribution $\mathbb{E}_{j \sim P_y}$ (described in next section). Because we view ProjE_pointwise as a multiclass classification problem, we use the sigmoid and tanh activation functions as our choice for $g(\cdot)$ and $f(\cdot)$ respectively. When deployed, the ranking score of the i$^\text{th}$ candidate-entity is:

$$h(\mathbf{e}, \mathbf{r})_i = \text{sigmoid}\left(\mathbf{W}^c_{[i,:]}\tanh\left(\mathbf{e} \oplus \mathbf{r}\right) + b_p\right), \tag{7}$$

where $\mathbf{W}^c_{[i,:]}$ represents $i^{th}$ candidate in the candidate-entity matrix.

Recently, softmax regression loss has achieved good results in multi-label image annotation tasks [12, 11]. This is because multi-label image annotation, as well as many other classification tasks, should consider their predicted scores collectively. Inspired by this way of thinking, we employ the softmax activation function in order to classify candidate-entities collectively, *i.e.*, using a listwise method. In this case we define the loss function of **ProjE_listwise** as:

$$\mathcal{L}(\mathbf{e}, \mathbf{r}, \mathbf{y}) = -\sum_i^{|\mathbf{y}|} \frac{\mathbb{1}(y_i = 1)}{\sum_i \mathbb{1}(y_i = 1)} \log\left(h(\mathbf{e}, \mathbf{r})_i\right), \tag{8}$$

where the target probability (*i.e.*, the target score) of a positive candidate is 1 / (total number of positive candidates of the input instance). Similar to Eq. 7, we replace $g(\cdot)$ and $f(\cdot)$ as softmax and tanh respectively, which can be written equivalently as:

$$h(\mathbf{e}, \mathbf{r})_i = \frac{\exp(\mathbf{W}^c_{[i,:]}\tanh(\mathbf{e} \oplus \mathbf{r}) + b_p)}{\sum_j \exp(\mathbf{W}^c_{[j,:]}\tanh(\mathbf{e} \oplus \mathbf{r}) + b_p)}. \tag{9}$$

Later, we perform a comprehensive set of experiments that compare ProjE with more than a dozen related models and discuss the proposed ProjE_pointwise and ProjE_listwise variants in depth.

## 2.3 Candidate Sampling

Although ProjE limits the number of additional parameters, the projection operation may be costly due to the large number of candidate-entities (*i.e.*, the number of rows in $\mathbf{W}^c$). If we reduce the number of candidate-entities in the training phrase, we could create a smaller working set that only contains a subset of the embedding matrix $\mathbf{W}^E$. With this in mind, we use candidate sampling to reduce the number of candidate-entities. Candidate sampling is not a new problem; many recent works have addressed this problem in interesting ways [16, 26, 13]. We experimented with many choices, and found that the negative sampling used in Word2Vec [26] resulted the best performance.

For a given entity $\mathbf{e}$, relationship $\mathbf{r}$, and a binary label vector $\mathbf{y}$, we compute the projection with all of the positive candidates and only a sampled subset of negative candidates from $\mathbf{P}_y$ following the convention of Word2Vec. For simplicity, $\mathbf{P}_y$ can be replaced by a $(0, 1)$ binomial distribution $B(1, p_y)$ shared by all training instances, where $p_y$ is the probability that a negative candidate is sampled and $1 - p_y$ is the probability that a negative candidate is not sampled. For every negative candidate in $\mathbf{y}$ we sample a value from $B(1, p_y)$ to determine whether we include this candidate in the candidate-entity matrix $\mathbf{W}^c$ or not.

In the Supplementary Material we evaluate the performance of ProjE with different candidate sampling rates $p_y \in \{5\%, 25\%, 50\%, 75\%, 95\%\}$. Our experiments show relatively consistent performance using negative sampling rates as low as $25\%$.

## 3 Experiments

We evaluate the ProjE model with entity prediction and relationship prediction tasks, and compare the performance against several existing methods using experimental procedures, datasets, and metrics established in the related work. The FB15K dataset is a 15,000-entity subset of Freebase; the Semantic MEDLINE Database (SemMedDB) is a KG extracted from all of PubMed [20]; and DBpedia is KG extracted from Wikipedia infoboxes [23]. Using DBpedia and SemMedDB, we also introduce a new fact checking task for a practical case study on the usefulness of these models. ProjE is implemented in Python using TensorFlow [1]; the code and data are available at `https://github.com/nddsg/ProjE`.

Table 2: Entity prediction on FB15K dataset. Missing values indicate scores not reported in the original work.

|  | Mean Rank | | HITS@10 (%) | |
| --- | --- | --- | --- | --- |
| Algorithm | Raw | Filtered | Raw | Filtered |
| Unstructured | 1074 | 979 | 4.5 | 6.3 |
| RESCAL | 828 | 683 | 28.4 | 44.1 |
| SE | 273 | 162 | 28.8 | 39.8 |
| SME (linear) | 274 | 154 | 30.7 | 40.8 |
| SME (bilinear) | 284 | 158 | 31.3 | 41.3 |
| LFM | 283 | 164 | 26.0 | 33.1 |
| TransE | 243 | 125 | 34.9 | 47.1 |
| DKRL (CNN) | 200 | 113 | 44.3 | 57.6 |
| TransH | 212 | 87 | 45.7 | 64.4 |
| TransR | 198 | 77 | 48.2 | 68.7 |
| TransE + Rev | 205 | 63 | 47.9 | 70.2 |
| HolE | - | - | - | 73.9 |
| PTransE (ADD, len-2 path) | 200 | 54 | 51.8 | 83.4 |
| PTransE (RNN, len-2 path) | 242 | 92 | 50.6 | 82.2 |
| PTransE (ADD, len-3 path) | 207 | 58 | 51.4 | 84.6 |
| TransA | 164 | 58 | - | - |
| **ProjE_pointwise** | 174 | 104 | **56.5** | 86.6 |
| **ProjE_listwise** | **146** | 76 | 54.6 | 71.2 |
| **ProjE_wlistwise** | **124** | **34** | 54.7 | **88.4** |

Table 3: Relationship prediction on FB15K dataset.

|  | Mean Rank | | HITS@1 (%) | |
| --- | --- | --- | --- | --- |
| Algorithm | Raw | Filtered | Raw | Filtered |
| TransE | 2.8 | 2.5 | 65.1 | 84.3 |
| TransE + Rev | 2.6 | 2.3 | 67.1 | 86.7 |
| DKRL (CNN) | 2.9 | 2.5 | 69.8 | 89.0 |
| PTransE (ADD, len-2 path) | 1.7 | **1.2** | 69.5 | 93.6 |
| PTransE (RNN, len-2 path) | 1.9 | 1.4 | 68.3 | 93.2 |
| PTransE (ADD, len-3 path) | 1.8 | 1.4 | 68.5 | 94.0 |
| **ProjE_pointwise** | 1.6 | 1.3 | 75.6 | 95.6 |
| **ProjE_listwise** | 1.5 | 1.2 | 75.8 | 95.7 |
| **ProjE_wlistwise** | 1.5 | 1.2 | 75.5 | 95.6 |

## 3.1 Settings

For both entity and relationship prediction tasks, we use Adam [21] as the stochastic optimizer with default hyper-parameter settings: $\beta_1 = 0.9$, $\beta_2 = 0.999$, and $\epsilon = 1e^{-8}$. During the training phrase, we apply an $L_1$ regularizer to all parameters in ProjE and a dropout layer on top of the combination operator to prevent over-fitting.

The hyper-parameters in ProjE are the learning rate $lr$, embedding size $k$, mini-batch size $b$, regularizer weight $\alpha$, dropout probability $p_d$, and success probability for negative candidate sampling $p_y$. We set $lr = 0.01$, $b = 200$, $\alpha = 1e^{-5}$, and $p_d = 0.5$ for both tasks, $k = 200$, $p_y = 0.5$ for the entity prediction task and $k = 100$, $p_y = 0.75$ for the relationship prediction task.

For all tasks, ProjE was trained for at most 100 iterations, and all parameters were initialized from a uniform distribution $U[-\frac{6}{\sqrt{k}}, \frac{6}{\sqrt{k}}]$ as suggested by TransE [4]. ProjE can also be initialized with pre-trained embeddings.

Table 4: AUC scores of fact checking test cases on DBpedia and SemMedDB.

|  | DBPedia | | | | | SemMedDB | |
| --- | --- | --- | --- | --- | --- | --- | --- |
| Algorithm | CapitalOf | Company CEO | NYT Bestseller | US Civil War | US Vice-President | Disease | Cell |
| Adamic/Adar | 0.387 | 0.665 | 0.650 | 0.642 | 0.795 | 0.671 | 0.755 |
| Semantic Proximity | 0.706 | 0.614 | 0.641 | 0.582 | 0.805 | 0.871 | 0.840 |
| SimRank | 0.553 | 0.824 | 0.695 | 0.685 | 0.912 | 0.809 | 0.749 |
| AMIE | 0.550 | 0.669 | 0.520 | 0.659 | 0.987 | 0.889 | 0.898 |
| PPR | 0.535 | 0.579 | 0.529 | 0.488 | 0.683 | 0.827 | 0.885 |
| PCRW | 0.550 | 0.542 | 0.486 | 0.488 | 0.672 | 0.911 | 0.765 |
| TransE | 0.655 | 0.728 | 0.601 | 0.612 | 0.520 | 0.532 | 0.620 |
| PredPath | 0.920 | 0.747 | 0.664 | 0.749 | 0.993 | **0.941** | 0.928 |
| **ProjE** | **0.979** | **0.845** | **0.852** | **0.824** | **1.000** | 0.926 | **0.971** |

## 3.2 Entity and Relationship Prediction

We evaluated ProjE's performance on entity and relationship prediction tasks using the FB15K dataset following the experiment settings in TransE [4] and PTransE [24]. For entity prediction, we aim to predict a missing $h$ (or $t$) for a given triple $\langle h, r, t \rangle$ by ranking all of the entities in the KG. To create a test set we replaced the head or tail-entity with all entities in the KG, and rank these replacement entities in descending order. For relationship prediction, we replaced the relationship of each test triple with all relationships in the KG, and rank these replacement relationships in descending order.

Following convention, we use mean rank and HITS@k as evaluation metrics. Mean rank measures the average rank of correct entities/relationships. HITS@k measures if correct entities/relationships appear within the top-$k$ elements. The filtered mean rank and filtered HITS@k ignore all other true entities/relationships in the result and only look at the target entity/relationship. For example, if the target relationship between ⟨Springfield, ?, Illinois⟩ is locatedIn, and the top-2 ranked relationships are capitalOf and locatedIn, then the raw mean rank and HITS@1 of this example would be 2 and 0 respectively, but the *filtered* mean rank and HITS@1 would both be 1 because the filtered mean rank and filtered HITS@k ignore the correct capitalOf relationship in the results set.

In addition to ProjE_pointwise and ProjE_listwise, we also evaluate **ProjE_wlistwise**, which is a slight variation of ProjE_listwise that incorporates instance-level weights ($\Sigma_i \mathbb{1}(y_i = 1)$) to increase the importance of N-to-N and N-to-1 (1-to-N) relationships.

Table 2 and Tab. 3 show that the three ProjE variants outperform existing methods in most cases. Table 3 contains fewer models than Tab. 2 because many models do not perform the relationship prediction task. We also adapt the pointwise and listwise ranking methods to TransE using the same hyperparameter settings, but the performance does not improve significantly and is not shown here. This indicates that the pointwise and listwise ranking methods are not merely simple tricks that can be added to any model to improve performance.

Surprisingly, although softmax is usually used in mutually exclusive multi-class classification problems and sigmoid is a more natural choice for non-exclusive cases like the KGC task, our results show that both ProjE_listwise and ProjE_wlistwise perform better than ProjE_pointwise in most cases.

This is because KGC is a special ranking task, where a good model ought to have the following properties: 1) the score of all positive candidates should be maximized and the score of all negative candidates should be minimized, and 2) the number of positive candidates that are ranked above negative candidates should be maximized. By maximizing the similarity between the ranking score vector and the binary label vector, ProjE_pointwise meets the first property but fails to meet the second, *i.e.*, ProjE_pointwise does not addresses the ranking order of all candidates collectively, because sigmoid is applied to each candidate individually. On the other hand, ProjE_listwise and ProjE_wlistwise successfully addresses both properties by maximizing the similarity between the binary label vector and the ranking score vector, which is an exponential-normalized ranking score vector that imposes an explicit ordering to the candidate-entities collectively.

In the Supplementary Material we also examine the stability of the proposed ProjE model and demonstrate that the performance of ProjE increases steadily and smoothly during training.

## 3.3 Fact Checking

Unlike the entity prediction and relationship prediction tasks that predict randomly sampled triples, we employ a new fact checking task that tests the predictive power of various models on real world questions. We view the fact checking task as a type of link prediction problem because a fact statement $\langle h, r, t \rangle$ can be naturally considered as an edge in a KG.

We use ProjE_wlistwise with a small change: rather than using entity embeddings directly, the input vector of ProjE consists of the *predicate paths* between the two entities [32]. We learn the entity-embeddings by adding an input layer that converts input predicate paths into the entity-embedding.

We employ the experimental setup and question set from Shi and Weninger (2016) on the DBPedia and SemMedDB data sets. Specifically, we remove all edges having the same label as the input relationship $r$ and perform fact checking on the modified KG by predicting the existence of $r$ on hundreds of variations of 7 types of questions. For example, the CapitalOf question checks various claims of the capitals of US states. In this case, we check if each of the 5 most populous cities within each state is its capital. This results in about $5 \times 5 = 250$ checked facts with an 20/80 positive to negative label ratio. The odds that some fact statement is true is equivalent to the odds that the fact's triple is missing from the KG (rather than purposefully omitted, *i.e.*, a true negative). Results in Tab. 4 show that ProjE outperforms existing fact checking and link prediction models [2, 6, 17, 9, 15, 22] in all but one question type.

# 4 Conclusions and Future Work

To recap, the contributions of the present work are as follows: 1) we view the KGC task as a ranking problem and project candidate-entities onto a vector representing a combined embedding of the known parts of an input triple and order the ranking score vector in descending order; 2) we show that by optimizing the ranking score vector collectively using the listwise ProjE variation, we can significantly improve prediction performance; 3) ProjE uses only directly connected, length-1 paths during training, and has a relatively simple 2-layer structure, yet outperforms complex models that have a richer parameter or feature set; and 4) unlike other models (*e.g.*, CVSM, RTransE, DKRL), the present work does not require any pre-trained embeddings and has many fewer parameters than related models. We finally show that ProjE can outperform existing methods on fact checking tasks.

For future work, we will adapt more complicated neural network models such RNN and CNN with the embedding projection model presented here. It is also possible to incorporate rich feature sets from length-2 and length-3 paths, but these would necessarily add additional complexity. Instead, we plan to use information from complex paths in the KG to clearly summarize the many complicated ways in which entities are connected.

# A  Appendix

In this supplement we provide a detailed algorithm description of the proposed ProjE_wlistwise model in Alg. 1, of which ProjE_listwise is a special case. Next, two more experiments are shown to demonstrate the training stability and scaling potential of ProjE.

## A.1  Training ProjE

In Alg. 1, we describe the training process of ProjE_wlistwise. For a given training triple set $\mathbf{S}$, we first construct the actual training set by randomly corrupting either the head entity $h$ or tail entity $t$, and then generate the corresponding positive and negative candidates from $\mathbf{S}$ using candidate sampling if requested. Then for each mini-batch in the newly generated training data set, we calculate the loss and update the parameters accordingly.

## A.2  Model Stability

In order to assess the training stability, we plotted the mean rank, filtered mean rank, HITS@10 and filtered HITS@10 over the first 25 training iterations on the FB15K dataset. For the purpose of illustration, we also draw three dashed lines representing the top-3 existing models that achieved the best performance in each metric.

As shown in Fig. 2, the performance of all three ProjE variants become stable after the first few iterations due to the use of Adam optimizer. The score variation between each iteration is also low, indicating stable training progress. The ProjE_wlistwise variant performed the best across all tests, followed by ProjE_listwise and ProjE_pointwise respectively.

**Input:** Training triples $\mathbf{S} = \{(h, r, t)\}$, entities $\mathbf{E}$, relations $\mathbf{R}$, embedding dimension $k$, dropout probability $p_d$, candidate sampling rate $p_y$, regularizer parameter $\alpha$.
**initialize** embedding matrices $\mathbf{W^E}, \mathbf{W^R}$, combination operators (diagonal matrices) $\mathbf{D}_{eh}, \mathbf{D}_{rh}, \mathbf{D}_{et}, \mathbf{D}_{rt}$ with uniform$(-\frac{6}{\sqrt{k}}, \frac{6}{\sqrt{k}})$
**Loop** /* A training iteration/epoch                                              */
    $\mathbf{S^h} \leftarrow \{\}, \mathbf{T^h} \leftarrow \{\}, \mathbf{S^t} \leftarrow \{\}, \mathbf{T^t} \leftarrow \{\}$; /* training data                        */
    **for** $(h, r, t) \in \mathbf{S}$ **do** /* construct training data using all training triples                */
        $e \leftarrow \text{random}(h, t)$;
        **if** $e == h$ **then** /* tail is missing                                             */
            $\mathbf{S^h}.\text{add}([e, r])$;
            /* all positive tails from $\mathbf{S}$ and some sampled negative candidates                        */
            $\mathbf{T^h}.\text{add}(\{t' | (h, r, t') \in \mathbf{S}\} \cup \text{sample}(\mathbf{E}, p_y))$;
        **else** /* head is missing                                                     */
            $\mathbf{S^t}.\text{add}([e, r])$;
            /* all positive heads from $\mathbf{S}$ and some sampled negative candidates                        */
            $\mathbf{T^t}.\text{add}(\{h' | (h', r, t) \in \mathbf{S}\} \cup \text{sample}(\mathbf{E}, p_y))$;
        **end**
    **end**
    **for** *each* $(\mathbf{S^h_b}, \mathbf{T^h_b}, \mathbf{S^t_b}, \mathbf{T^t_b}) \subset (\mathbf{S^h}, \mathbf{T^h}, \mathbf{S^t}, \mathbf{T^t})$ **do** /* mini-batches          */
        $l \leftarrow 0$;
        **for** $(\mathbf{s}^h, \mathbf{t}^h, \mathbf{s}^t, \mathbf{t}^t) \in (\mathbf{S^h_b}, \mathbf{T^h_b}, \mathbf{S^t_b}, \mathbf{T^t_b})$ **do** /* training instance           */
            $o_h \leftarrow \text{softmax}(\mathbf{W}^E_{[\mathbf{t}^t,:]} \times \tanh(\text{dropout}(p_d, \mathbf{D}_{et} \times (\mathbf{W}^\mathbf{E}_{[\mathbf{s}^t[0],:]})^T + \mathbf{D}_{rt} \times (\mathbf{W}^\mathbf{R}_{[\mathbf{s}^t[1],:]})^T + \mathbf{b_c})) + b_p)$;
            $o_t \leftarrow \text{softmax}(\mathbf{W}^E_{[\mathbf{t}^h,:]} \times \tanh(\text{dropout}(p_d, \mathbf{D}_{eh} \times (\mathbf{W}^\mathbf{E}_{[\mathbf{s}^h[0],:]})^T + \mathbf{D}_{rh} \times (\mathbf{W}^\mathbf{R}_{[\mathbf{s}^h[1],:]})^T + \mathbf{b_c})) + b_p)$;
            $l = l - \Sigma(\{\mathbb{1}((h, \mathbf{s}^t[1], \mathbf{s}^t[0]) \in \mathbf{S}) | h \in \mathbf{t}^t\} \circ \log(o_h)) - \Sigma(\{\mathbb{1}((\mathbf{s}^h[0], \mathbf{s}^h[1], t) \in \mathbf{S}) | t \in \mathbf{t}^h\} \circ \log(o_t))$;
        **end**
        /* L1 loss                                                                */
        $l_r \leftarrow \text{Regu}_1(\mathbf{W^E}) + \text{Regu}_1(\mathbf{W^R}) + \text{Regu}_1(\mathbf{D}_{eh}) + \text{Regu}_1(\mathbf{D}_{rh}) + \text{Regu}_1(\mathbf{D}_{et}) + \text{Regu}_1(\mathbf{D}_{rt})$;
        update all parameters w.r.t. $l + \alpha l_r$;
    **end**
**EndLoop**

**Algorithm 1:** Algorithm of ProjE_wlistwise Training. $\circ$ is Hadamard product and $\times$ is matrix product.

## A.3 Candidate Sampling

In order to evaluate the relationship between the sampling rate and the model performance, we plotted five different $p_y$ rates from $5\%$ to $95\%$ using the ProjE_wlistwise variant. All settings except $p_y = 5\%$ achieved better performance than the top-3 existing methods in each metric. These results demonstrate that that we can use ProjE with a relatively small sampling rate ($25\%$), but it also demonstrates that ProjE is robust in the presence of different positive-to-negative training data ratios. Indeed, we find that the best results are often achieved under the 25% sampling ratio. This robustness provides ProjE the ability to handle very large datasets by significantly reducing the active working set.

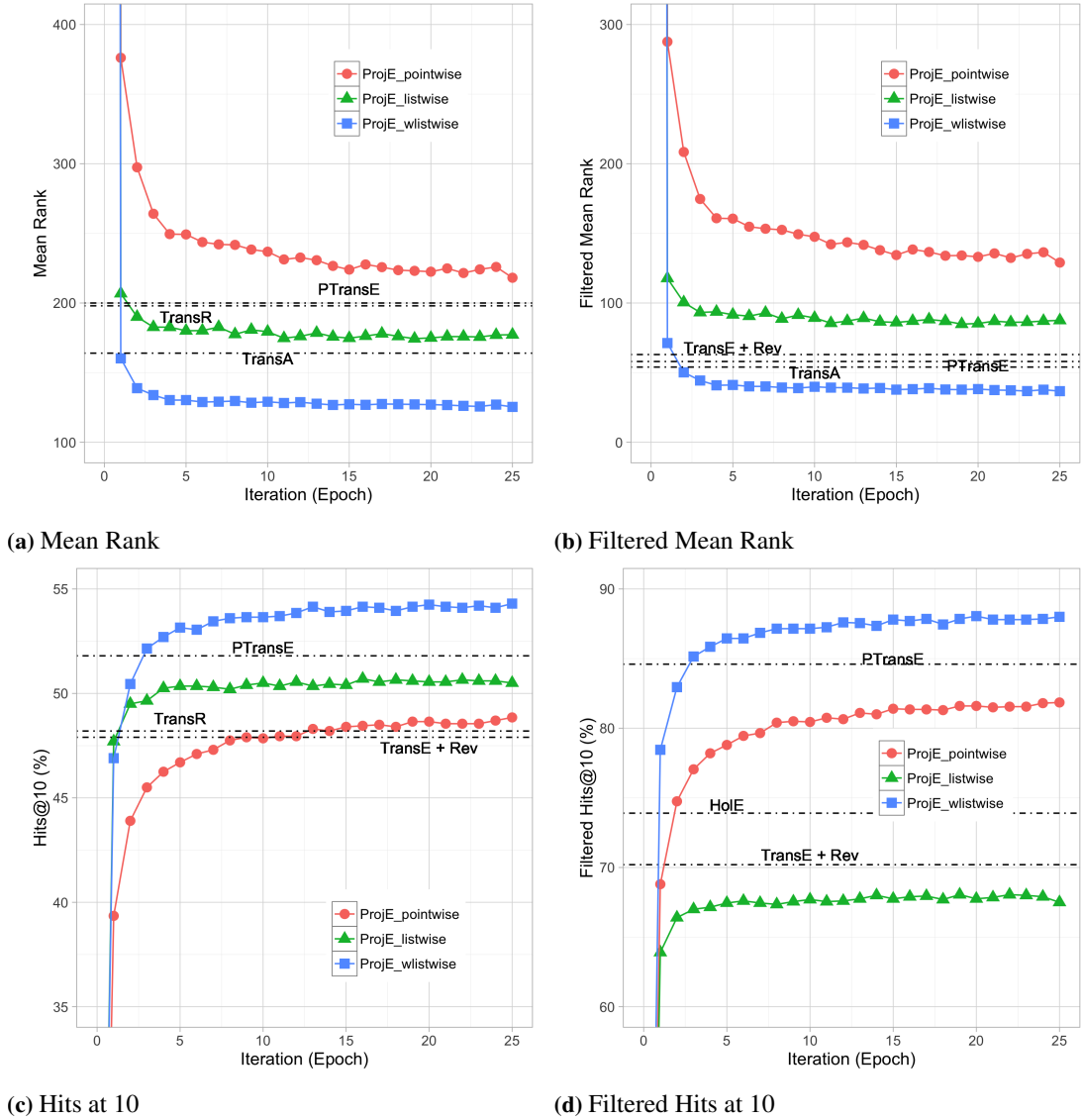

(a) Mean Rank

(b) Filtered Mean Rank

(c) Hits at 10

(d) Filtered Hits at 10

**Figure 2:** ProjE variants on the FB15K dataset. Each plot contains three dashed lines representing the top-3 existing models that achieved the best performance in each metric.

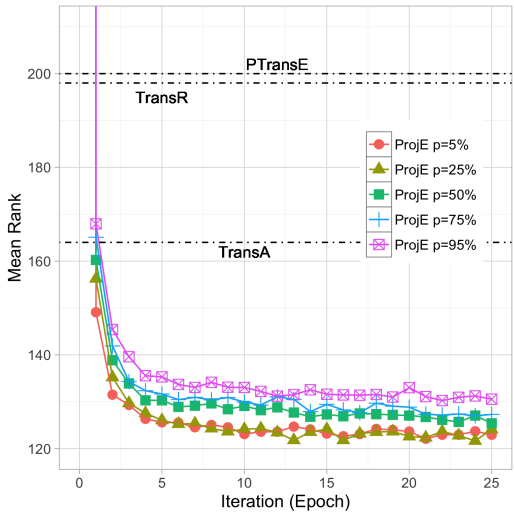

(a) Mean Rank

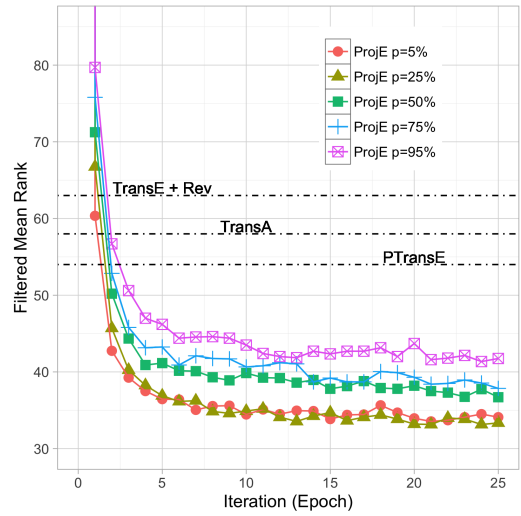

(b) Filtered Mean Rank

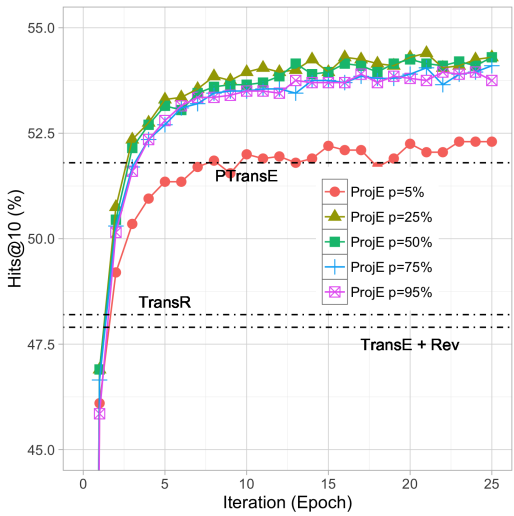

(c) Hits at 10

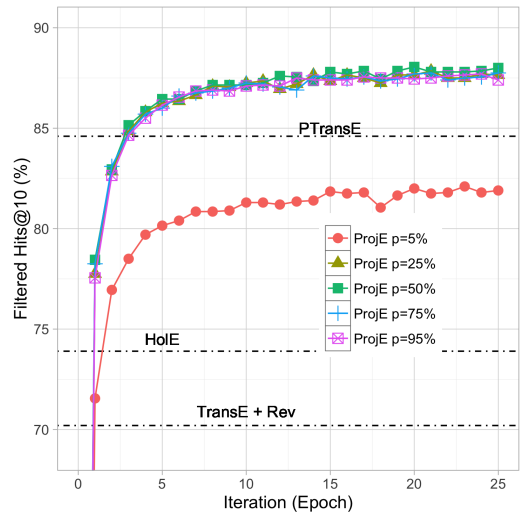

(d) Filtered Hits at 10

**Figure 3:** ProjE_wlistwise with different candidate sampling $p_y$ rate on FB15K. Each plot contains three dashed lines representing the top-3 existing models that achieved the best performance in each metric.